\documentclass[a4paper,conference]{IEEEtran}
\IEEEoverridecommandlockouts

\usepackage{cite}
\usepackage{amsmath,amssymb,amsfonts}
\usepackage{graphicx}
\usepackage{booktabs}
\usepackage{url}
\usepackage{algorithmic}
\usepackage{subcaption}
\usepackage[normalem]{ulem}

\newcommand{\authorfont}{\fontsize{11pt}{10pt}\selectfont}
\renewcommand\IEEEkeywordsname{Keywords}

\usepackage[T1]{fontenc}
\usepackage[latin1, utf8]{inputenc}
\usepackage{babel}
\usepackage{booktabs}
\usepackage[table]{xcolor}
\usepackage{hhline}
\usepackage{pgf}
\usepackage{multirow,multicol}
\usepackage[hidelinks]{hyperref}
\usepackage{amsmath,bm}

\usepackage{collcell}
\usepackage{bbding} 
\definecolor{Gray}{gray}{0.9}

\begin{document}
\bstctlcite{IEEEexample:BSTcontrol}



\title{Speech and Natural Language Processing Technologies for Pseudo-Pilot Simulator}

\author{\IEEEauthorblockN{\authorfont Amrutha Prasad$^{\star,\dagger,1,2}$, Juan Zuluaga-Gomez$^{\dagger,1,3}$, Petr Motlicek$^{1,2}$, Saeed Sarfjoo$^{1}$, \\ Iuliia Nigmatulina$^{1,4}$, Karel Vesely$^{2}$}
\IEEEauthorblockA{
\\ \normalsize $^{1}$Idiap Research Institute, Martigny, Switzerland\\
\normalsize $^{2}$Brno University of Technology, Speech@FIT, IT4I CoE, Brno, Czech Republic \\
\normalsize $^{3}$Ecole Polytechnique Federale de Lausanne (EPFL), Lausanne, Switzerland \\
\normalsize $^{4}$Institute of Computational Linguistics, University of Zurich, Switzerland \\
\normalsize $^\star$corresponding author: amrutha.prasad@idiap.ch \\
\normalsize $^\dagger$equal contribution}
}

\maketitle

\begin{abstract}
This paper describes a simple yet efficient repetition-based modular system for speeding up air-traffic controllers (ATCos) training. E.g., a human pilot is still required in EUROCONTROL's ESCAPE lite simulator~\texttt{\url{https://www.eurocontrol.int/simulator/escape}} during ATCo training. However, this need can be substituted by an automatic system  that could act as a pilot. In this paper, we aim to develop and integrate a pseudo-pilot agent into the ATCo training pipeline by merging diverse artificial intelligence (AI) powered modules. The system understands the voice communications issued by the ATCo, and, in turn, it generates a spoken prompt that follows the pilot's phraseology to the initial communication. Our system mainly relies on open-source AI tools and air traffic control (ATC) databases, thus, proving its simplicity and ease of replicability. 
The overall pipeline is composed of the following: 
(1) a submodule that receives and pre-processes the input stream of raw audio, (2) an automatic speech recognition (ASR) system that transforms audio into a sequence of words; 
(3) a high-level ATC-related entity parser, which extracts relevant information from the communication, i.e., callsigns and commands, 
and finally, (4) a speech synthesizer submodule that generates responses based on the high-level ATC entities previously extracted. 
Overall, we show that this system could pave the way toward developing a real proof-of-concept pseudo-pilot system. Hence, speeding up the training of ATCos while drastically reducing its overall cost. 
\end{abstract}


\section{Introduction}

The communication between air-traffic controllers (ATCos) and pilots often requires the  ATCos to issue commands for the safe travel of an aircraft. Although different means of communication such as CPDLC\footnote{Controller Pilot Data Link Communications (CPDLC). CPDLC is a two-way data-link system by which controllers can transmit non-urgent strategic messages to an aircraft as an alternative to voice communications. The message is displayed on a flight deck visual display.}
messages, flight strips, or shortcodes are used, voice is the central part of these communications. Recent research has focused on improving the assistant-based speech recognition system (ABSR) for ATCos to reduce overall workload while increasing safety~\cite{helmke2016reducing,ohneiser2021robust}. Another application in which speech recognition can be used in is the training of the ATCos.  
Training air-traffic controllers (ATCos) usually require a human pseudo-pilot.
The pseudo-pilots respond or issues requests to the ATCo trainee to simulate a standard ATC communication scenario for an ATCo.
The ATCo training is a human-intensive task, specialized workforce, and the overall cost is usually high. 
The pseudo-pilots are required to follow the commands issued by the ATCos. Also, they update the simulator so the ATCos can see if the pilots follow the orders similar to the actual situation. 

\begin{figure*}[t!]
    \centering
    \includegraphics[width=0.95\linewidth]{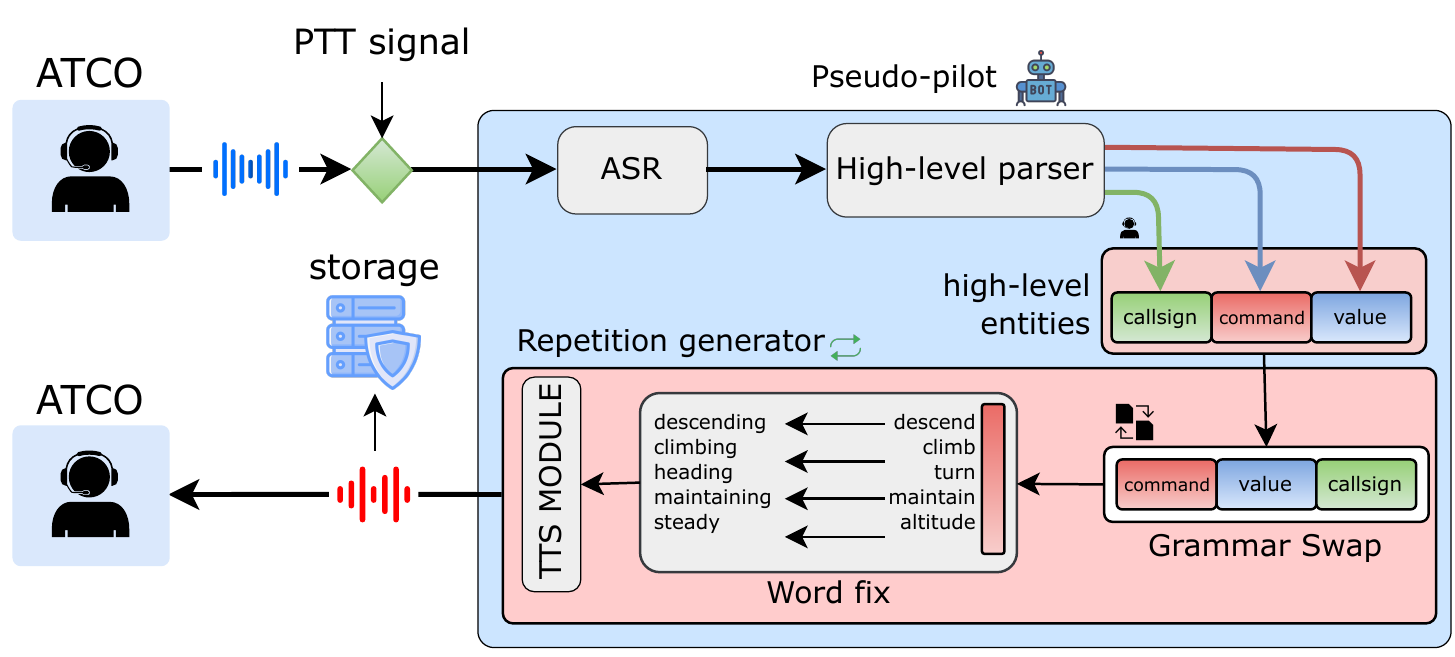}
    \caption{\textbf{Overall pipeline for simulating a pilot for air traffic controller training.} The pipeline starts with an ATCo issuing a communication and its capture after the end of the PTT signal. Then, the speech-to-text and high-level entity parser modules transcribe and extract the ATC-related entities from the voice communication. The output is then accustomed to `pilots' structured grammar. The speech synthesizer then uses the generated text to create a WAV file containing the spoken textual prompt. In the end, a pseudo-pilot response is produced that can provide pertinent feedback to ATCo trainees.}
    \label{fig:full_pipeline}
\end{figure*}

An autonomous pilot agent could aid different parts of ATCo's training process to reduce the training costs due to specialized human involvement. For instance, previous research~\cite{lin2021deep,zhang2022automatic} has investigated deep learning based framework for only repetition generator.
In this work, we develop a modular, simple, and easy-to-deploy artificial intelligence (AI) based system for simulating a pilot, i.e., a pseudo-pilot. The main focus is to develop a pipeline from transcribing a controller's speech to generating the entities, which produces a response to simulate a pilot.
Figure~\ref{fig:full_pipeline} describes the overall pipeline used in our system. The first module uses an automatic speech recognition (ASR) system to convert the voice communication issued by the ATCo trainee into text. The following module is a Named Entity Recognition (NER) system\footnote{We also refer to it as `high-level entity parser' in several parts of the paper.} that maps the text to different ATC-related categories such as callsigns, commands, and values.
The last module is a repetition generator that returns the pilot response based on ATCo's voice message. This module consists of two submodules: (i) a rule-based system to convert the text into an appropriate pilot response, and (ii) a text-to-speech (TTS) system\footnote{Also known as a speech synthesizer system.} that generates the corresponding response into a spoken form, i.e., speech play in the headphones of the ATCO trainee.

The proposed system handles the most frequent commands in ATC communications (see a list of commands and words in Figure~\ref{fig:full_pipeline}). In summary, the proposed pseudo-pilot agent is well suited for the early stages of ATCo training, where potential controllers are required to build fundamental knowledge and skills.  
The proposed system can be adapted easily to specific conditions based on different airports where the spoken language is foreign. 
Finally, our system is versatile due to its simplicity and modularity nature. For example, we can plug in or out different modules depending on the need of the specific training setup.




The rest of the paper is organized as follows.
Section~\ref{sec:related_work} briefly describes previous research in the field of ASR, NER and TTS systems. This is followed by Section~\ref{sec:datasets} that describes the data used in this work for training and evaluating different systems.
Section~\ref{sec:training-system} describes the details of the modules used in this research and presents experimental results for ASR and NER systems. This is followed by conclusions in Section~\ref{sec:conclusion}.


\section{Related Work}
\label{sec:related_work}

\subsection{Air-traffic Controller Training System}

Air traffic control communications between ATCos and pilots are crucial for the safe travel of an aircraft. The prominent part of this communication comes from the ATCos. 
They issue commands from tens to hundreds of aircraft in a short period, and this number varies depending on the airspace condition and period of the year (e.g., ATC operations shows increased workload during summer for ATCos).
Hence, ATCo trainees must learn how to handle stressful and complex airspace situations. They also undergo different training procedures, for instance, communication with pilots, and this often happens in a simulation environment with human pseudo-pilots. 

As previously covered in~\cite{doc10056}, the ATCos have 3 stages of training: (i) initial, (ii) operational, and (iii) continuation training. As the training of controllers is a crucial component in air-traffic management (ATM), efforts have been made to develop simulation devices for the same~\cite{pavlinovic2017air,jurivcic2011role,chhaya2018enhancing}. Research also focused on the post-evaluation of this simulation-based training~\cite{eide2014post,nemethova2019education} and optimizing the training procedure~\cite{updegrove2017optimization} which still includes the involvement of human pseudo-pilot and the cost of training is still high.
However, in the recent works of ~\cite{lin2021deep,zhang2022automatic}, the authors develop a deep learning-based framework for the repetition generator to implement an autonomous pilot agent (APA). In~\cite{lin2021deep}, the authors develop APA with the main focus on repetition generator and text-to-speech system. The authors used a sequence-to-sequence mapping for the repetition generator and a transformer model~\cite{vaswani2017attention} to generate speech.


\subsection{Automatic Speech Recognition}

Automatic Speech Recognition (ASR) also termed a speech-to-text system converts speech in a given language to text. A standard ASR system employs an acoustic model (AM) and a language model (LM) to achieve this task. The former, AM, is trained with a set of speech recordings with a corresponding text, also referred to as transcripts. The AM represents the relationship between a speech signal and phonemes, or other linguistic units, that make up the speech. The latter, LM, is trained on a large corpus of text data. The LM is usually represented by a probability distribution over sequences of words. The LM provides context to distinguish between words and phrases that sound similar. Using the knowledge of AM and LM, a decoding graph is usually built as a Weighted Finite State Transducer (WFST)~\cite{mohri2002weighted,mohri2008speech,riley2009openfst}. The WFST graph generates text output given an observation sequence as shown in Figure~\ref{fig:asr_hybrid}.

Generally, recorded speech is represented as a sequence of acoustic feature vectors or \textit{observations}, $\bm{X}$; and the output word sequence, $\bm{W}$. During recognition or \textit{decoding}, the main goal is finding the most likely $\bm{W}$ given the input sequence $\bm{X}$. To solve this task, statistical models are trained using a \textit{corpus} $\bm{\mathcal{D}}$ of labelled training \textit{utterances}, ($\bm{X^n}$, $\bm{W^n}$). In general, if $\bm{X}$ is the sequence of acoustic features and $\bm{W}$ denotes a word sequence, the most likely word sequence is given by $\bm{\hat{W}}$:

\begin{equation}
    \bm{\hat{W}}=arg~\underset{\bm{W}}max~\bm{P} (\bm{W} | \bm{X}).
    \label{eq:asr}
\end{equation}

\noindent The problem is further reformulated using Bayes Theorem:

\begin{equation}
    \bm{P}(\bm{W}|\bm{X}) = \dfrac{p(\bm{X}|\bm{W})p(\bm{W})}{p(\bm{X})},
    \label{eq:asr_1}
\end{equation}

where, $\bm{P(X|W)}$ stands for the likelihood of the feature sequence $\bm{X}$, given the hypothesized word sequence, $\bm{W}$. $P(\bm{W})$ is the probability of the word sequence (normally, computed from a pretrained LM). $P(\bm{X})$ is the a-priori probability of the feature sequence $\bm{X}$, but it is ignored during the maximization operation due to its non-dependency to the AM and LM. Equation \ref{eq:asr_fin} is further simplified as $P(\bm{X})$ is a constant for any word sequence, as follows:

\begin{equation}
    \bm{\hat{W}}=arg~\underset{\bm{W}}max~p(\bm{X}|\bm{W})~p(\bm{W}).
    \label{eq:asr_fin}
\end{equation}

Standard ASR systems rely on a lexicon, language, and acoustic model as stated above. Currently, there are two main ASR paradigms, where different strategies, architectures, and procedures are employed for blending all these modules in one “system”.

\subsubsection{Hybrid Based ASR}
\label{subsec:hb-asr}

Automatic speech recognition with hybrid systems is based on Hidden Markov Models (HMM) and Deep Neural Networks (DNN)~\cite{dahl2011context,vesely2013sequence}. DNNs are an effective module for the estimating the posterior probability of a given set of possible outputs (e.g., phone-state or tri-phone state probability estimator, in ASR systems). These posterior probabilities can be seen as pseudo-likelihoods or “scale likelihoods”, which can be interfaced with HMM modules. HMMs provide a structure for mapping a temporal sequence of acoustic features, $\bm{X}$, e.g., Mel-frequency cepstral coefficients (MFCCs) into a sequence of states \cite{morgan1993hybrid,bourlard1993connectionist}. Hybrid systems remain one of the best approaches for building ASR engines. Currently, HMM-DNNs based ASR is the state-of-the-art systems for ASR in ATC domain~\cite{srinivasamurthy2017semi,kleinert2018semi,zuluaga2020automatic}. 

Moreover, recent work in ASR has targeted different areas in ATC. For instance, a benchmark for ASR on ATC communications databases is established in~\cite{zuluagagomez20_interspeech}. Leveraging non-transcribed ATC audio data by semi-supervised learning has been covered in~\cite{srinivasamurthy2017semi,kleinert2018semi,zuluagagomez21_interspeech}. Previous work related to the large-scale automatic collection of ATC audio data from different airports worldwide is covered in~\cite{kocour2021automatic,zuluaga2020automatic}. Additionally, innovative research targeted to improve callsign recognition by integrating surveillance data into the pipeline is covered by~\cite{kocour21_interspeech,nigmatulina2021improving,nigmatulina2022two}.

\begin{figure}[t]
    \centering
    \includegraphics[width=0.99\linewidth]{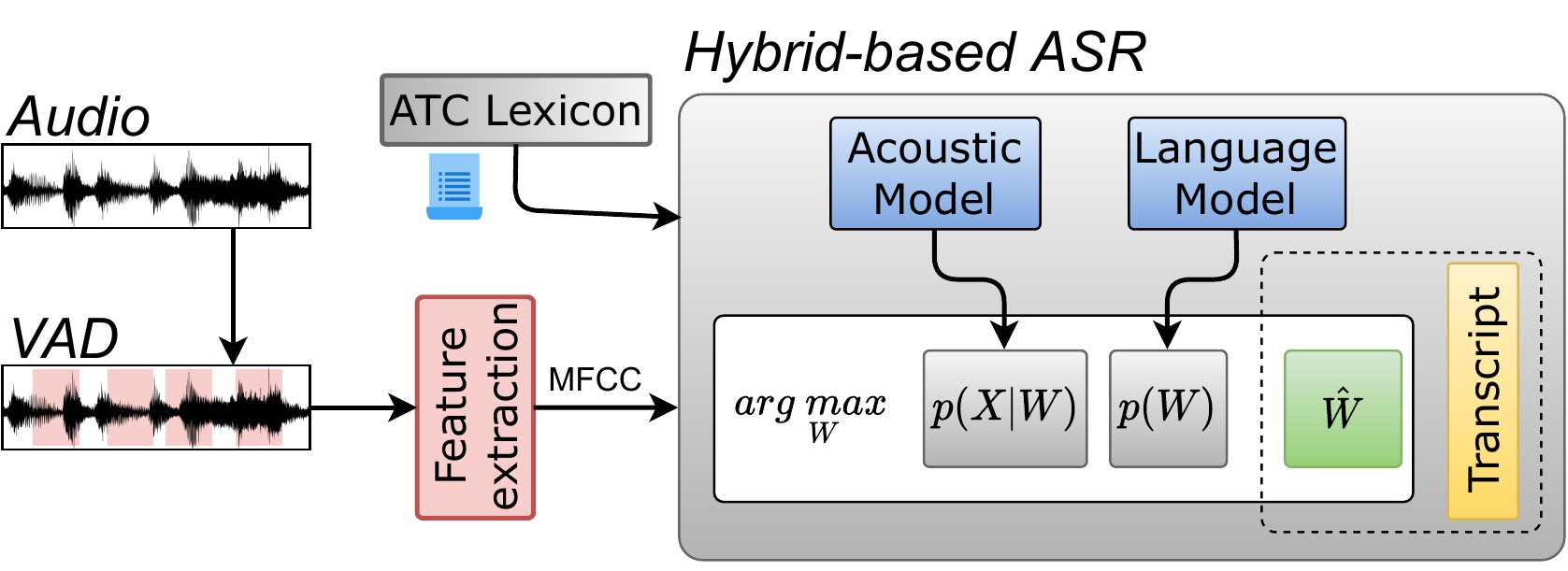}
    \caption{\textbf{Traditional automatic speech recognition system based on hidden Markov models and deep neural networks}. The system receives as input an ATCo voice communication, and it produces transcripts as output.}
    \label{fig:asr_hybrid}
\end{figure}

As depicted in Figure~\ref{fig:asr_hybrid}, the main components of a hybrid system are: a pronunciation lexicon, language model, and acoustic model. 
One key advantage of a hybrid system versus other ASR techniques is that the text data (e.g., words, dictionary) and pronunciation of new words are collected and added beforehand, hoping to match the target domain of the recognizer. Standard hybrid-based ASR approaches still rely on word-based lexicons, i.e., missing or out-of-vocabulary words from the lexicon cannot be hypothesized by the ASR decoder. 



\subsubsection{End-to-End ASR}

End-to-end (E2E) speech recognition is another paradigm for performing ASR. E2E-ASR aims at directly transcribing speech to text without requiring alignments between acoustic frames (i.e., input features) and output characters/words, which is normally required in hybrid-based ASR (see Section~\ref{subsec:hb-asr}). 
Unlike the hybrid approaches, the E2E model learns a direct mapping between acoustic frames and character units or words in one step towards the final objective of interest.
    

Recent work on encoder-decoder ASR can be categorized in two main approaches: Connectionist Temporal Classification (CTC)~\cite{graves2014towards} and attention-based encoder-decoder systems ~\cite{chorowski2015attention}. CTC uses intermediate label representation, allowing repetitions of labels and occurrences of the so-called `blank output' to label an output with `no label'. Attention-based encoders-decoders directly learn a mapping from input acoustic frames to character sequences. At each output time step, the model emits a character unit conditioned on the inputs and the history of the produced outputs. 
Related work on self-supervised learning~\cite{schneider2019wav2vec} for speech representation covers bidirectional models~\cite{baevski2020wav2vec,chen2021wavlm} and autoregressive models~\cite{oord2018representation, baevski2019vq}. 
An innovative research on E2E-ASR for ATC domain is covered in~\cite{zuluaga2022does} in which the authors fine-tuned a Wav2Vec~2.0 model~\cite{baevski2020wav2vec} with public and private ATC databases, reaching on-par performances with hybrid-based ASR models.

\begin{figure*}[!t]
    \centering
    \includegraphics[width=0.9\linewidth]{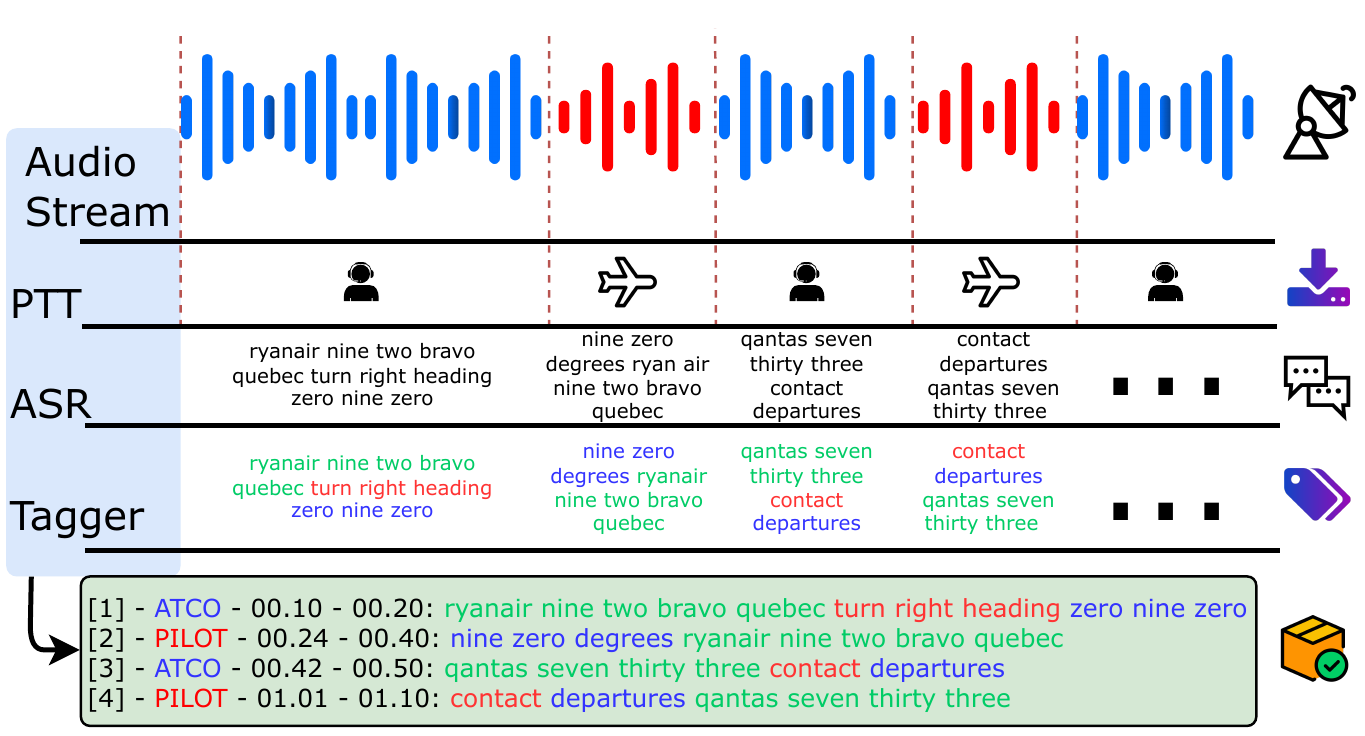}
    \caption{\textbf{Detailed outputs of the main ML-based submodules of our proposed pseudo-pilot system}. It includes pre-processing from the input audio stream, speaker role detection by push-to-talk (PTT) signal, transcripts generation and high-level ATC entities extraction with our speech-to-text and NER modules, respectively. All the data is later aggregated, packaged and sent to the repetition generator and TTS module. Note that this data can also be logged into a database for control and record.}
    \label{fig:metadata_pipeline}
\end{figure*}

\subsection{High-Level Entity Parser}

A high-level entity parser\footnote{In a more technical domain, for instance, natural language processing, the task above-mentioned is called named-entity recognition (NER). For simplicity and because it is more aligned to ATC domain, we term it \mbox{`high-level entity parsing'}.} system has the task to detect, classify and extract keywords from a given snippet of text or transcribed ATC communication. These keywords normally fall into certain pre-defined categories such as parts of speech, location, organizations, or proper nouns like persons' names. Based on the domain, these entities can differ. In the field of ATC we have defined as `key entities': callsigns, commands, values, and units. For example, the following transcribed communication (example taken from Figure~\ref{fig:metadata_pipeline}):

\vspace{0.2cm}

\noindent \textbf{Input:} \textcolor{black}{\dashuline{ryanair nine two bravo quebec turn right heading zero nine zero}}, 

\vspace{0.2cm} 

\noindent would be parsed to high-level entity format: 

\vspace{0.2cm} 

\noindent \textbf{Output:} \textcolor{teal}{\dashuline{<callsign> ryanair nine two bravo quebec </callsign>}} 
\textcolor{red}{\dashuline{<command> turn right heading </command> }} 
\textcolor{blue}{\dashuline{<value> zero nine zero </value>}} 
 
\vspace{0.2cm}

Early research~\cite{grishman1996message} on high-level entity parsing or `NER', used to obtain these tags from handcrafted dictionaries and ontologies. This, in turn, increased the overall complexity and was prone to human errors when escaling up to more entities or when adapting the system to a different domain. Collobert et al.~\cite{collobert2011natural} introduced machine learning-based methods for text processing in topics such as part-of-speech tagging, chunking, NER, and semantic role labeling. Further work on NER was carried by~\cite{piskorski2017first,yadav2018survey}. In practice, a high-level entity parser system can be developed by fine-tuning a pre-trained LM on the downstream NER task. State-of-the-art NER systems are based on well-known pretrained LMs such as, BERT~\cite{devlin-etal-2019-bert}, RoBERTa~\cite{liu2019roberta}, or DeBERTa~\cite{he2021deberta}. In our experiments, we adopted one of the most well-known and simple to use LM, i.e., BERT.


\subsection{Speech Synthesis}
\label{subsec:tts}

Speech synthesis also known as text-to-speech (TTS) is a technology involving several research fields such as linguistics, acoustics, speech signal processing, among others. TTS aims at transforming input text into a speech signal. There have been several approaches in the framework of TTS such as, formant-based parametric synthesis~\cite{klatt1987review}, waveform concatenation~\cite{murray1996emotional}, or SPSS-based\footnote{SPSS stands for statistical parametric speech synthesis.} models~\cite{tokuda2013speech}. Akin to language modeling, recent advances in deep learning has also impacted TTS. For instance, the widely known Tacotron model~\cite{Wang2017tacotron} was proposed in 2017 and Tacotron2~\cite{shen2018tacotron2}, an updated version, in 2018. These models are end-to-end generative TTS systems capable to synthesize speech from characters (or words). More recently, FastSpeech2~\cite{ren2020fastspeech} has gained a lot of popularity in the TTS research field due to its simplicity and because it works in a non-autoregressive manner. We refer the reader to the references cited above to get a more technical background behind these state-of-the-art end-to-end TTS systems.

\section{Datasets}
\label{sec:datasets}

This section briefly describes the public and private databases used for training and evaluating the ASR and high-level entity parser modules. Table~\ref{tab:databases} lists a summary of the databases employed in this research. It also lists the amount of train/test samples and their open-source status.

\begin{table}[t]
    \caption{\textbf{Air traffic control communications-related databases used for developing our systems.} $^\dagger$total number of hours of audio after silence removal.}
    \label{tab:databases}
    \centering
    \begin{tabular}{lcccccc}
        \toprule
        \rowcolor{Gray} \textbf{Database} &  & \multicolumn{2}{c}{\textbf{Duration}$^\dagger$} &  & \textbf{Open} & \textbf{Ref} \\
        \cline{3-4}
        \rule{0pt}{2ex}  &  & Train & Test &  & \textbf{source}  &  \\
        \midrule
        \rowcolor{Gray} \multicolumn{7}{c}{\textbf{Private databases}} \\
        \midrule
        HAAWAII &  & 43 & 4 &  & \XSolidBrush &  \cite{zuluaga2022does} \\
        Internal Data &  & 95 & - &  & \XSolidBrush & \cite{zuluagagomez20_interspeech}\\
        \midrule
        \rowcolor{Gray} \multicolumn{7}{c}{\textbf{Public databases}} \\
        \midrule
        ATCOSIM &  & 8 & - &  & \checkmark & \cite{ATCOSIM} \\
        UWB-ATCC &  & 10.4 & - &  & \checkmark & \cite{PILSEN_ATC} \\
        LDC-ATCC &  & 23 & 2.6 &  & \checkmark & \cite{LDC_ATCC} \\
        ATCO2-PL &  & 100 & 4 &  & \checkmark & \cite{kocour2021automatic} \\
        \bottomrule
    \end{tabular}
\end{table}

\textbf{LDC-ATCC corpus:} the Air Traffic Control Corpus\footnote{LDC-ATCC: \texttt{\url{https://catalog.ldc.upenn.edu/LDC94S14A}}.} (ATCC) consists of recorded speech for use in ATC research in the area of ASR and NLP. The audio data contains voice communication traffic between various ATCos and pilots. The audio files are sampled at 8 kHz,
16-bit linear, representing continuous monitoring without squelch or silence elimination. Each file has a single frequency over one to two hours of audio. The corpus contains gold annotations and metadata.\footnote{Metadata covers voice activity segmentation details, speaker role information (who is talking), and callsigns in ICAO format.}
The corpus consists of approximately 70\,h and after silence removal, the total duration of data is around 25\,h of ATCo and pilot transmissions.

\textbf{UWB-ATCC corpus:} the UWB-ATCC\footnote{Corpus released by the University of West Bohemia: \texttt{\url{https://lindat.mff.cuni.cz/repository/xmlui/handle/11858/00-097C-0000-0001-CCA1-0}}.} corpus is a free and public resource for research on ATC. It contains recordings of communication between ATCos and pilots. The speech is manually transcribed and labeled with the speaker information, i.e., pilot/controller. The total amount of speech after removing silences is 13\,hrs. The audio data is mono-channel sampled at 8kHz and 16-bit PCM.

\textbf{ATCO2 corpus:} dataset built for the development and evaluation of ASR and NLP technologies for English ATC communications. The dataset consists of English coming from LKTB, LKPR, LZIB, LSGS, LSZH, LSZB and YSSY airports. There are two official partitions, namely, \textit{ATCO2 test set 1h corpus} and the \textit{ATCO2 test set corpus}. The first corpus contains 1.1\,hr of open-sourced transcribed annotations, and it can be accessed for free in \texttt{\url{https://www.atco2.org/data}}. The latter contains around 3\,hrs of extra annotated data, and the full corpus is available for purchase through ELDA in
\texttt{\url{http://catalog.elra.info/en-us/repository/browse/ELRA-S0484}}. The recordings of both corpus are mono-channel sampled at 16kHz and 16-bit PCM. 

\textbf{HAAWAII corpus}\footnote{Highly Advanced Air Traffic Controller Workstation with Artificial Intelligence Integration: \texttt{\url{https://www.haawaii.de}}.}: dataset based on an exploratory research project that aims to research and develop a reliable and adaptable solution to automatically transcribe voice commands issued by both ATCos and pilots. The controller and pilot conversations are obtained from two Air Navigation Service Providers (ANSPs): (i) NATS for London Approach and (ii) ISAVIA for Icelandic en-route. The total amount of manually transcribed data available is around 47\,h (partitioned into 43\,h for train and 4\,h for test). Similar to another corpus, the audio files are sampled at 8\,kHz and 16-bit PCM.

\begin{figure}[!t]
    \centering
    \includegraphics[width=0.8\linewidth]{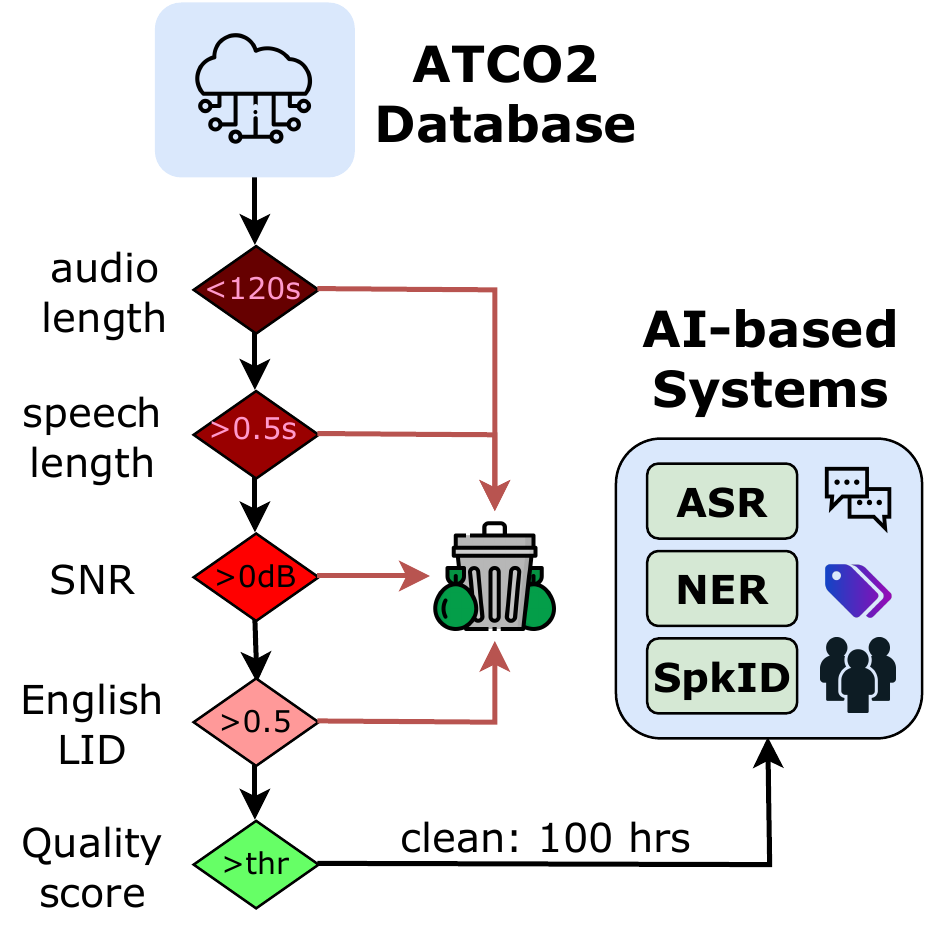}
    \caption{\textbf{Pipeline breakdown of data selection and filtering for training of the speech-to-text system}. SNR: signal-to-noise ratio; English LID: English language identification score. We apply these filters on top of the released version of \textit{ATCO2 pseudo labeled set corpus} available for purchase through ELDA in \texttt{\url{http://catalog.elra.info/en-us/repository/browse/ELRA-S0484}}.}
    \label{fig:data_flow}
\end{figure}

\section{Air-traffic Controller Training System}
\label{sec:training-system}

In this section, we describe the core modules 
and present the obtained results
of the proposed pseudo-pilot system. The first part covers pre-processing, automatic speech recognition, and high-level ATC-entity parsing, as described in Figure~\ref{fig:metadata_pipeline}. The final output of these modules then creates the spoken sentence using the speech synthesizer to simulate a pilot (second part). Figure~\ref{fig:full_pipeline} gives a broad overview of the system.

\subsection{Pre-processing}
\label{subsec:pre-processing}

As mentioned in Section~\ref{sec:datasets}, the sampling rate of all databases are not the same. Thus, the audio data is first up/down sampled to 16kHz. For ATCOs' data, the following steps are applied to select the best audio data from the open-source \textit{ATCO2 PL set corpus}\footnote{The \textit{ATCO2 PL set corpus} is composed of generated ASR transcripts, i.e., it is not manually annotated or corroborated by a human.}, which is described in Figure~\ref{fig:data_flow}. 
Firstly, the available data is filtered to remove very long (>120\,s) and short recordings (<0.5\,s). Next, we remove segments with SNR below 0 dB in order to have reasonably good audio quality. As described previously, the ATCO2 corpus consists of data from different airports, and thus, there are some non-English recordings. However, our ASR and NLP systems are trained with only English data. Thus, in the next step, we apply English language detection and remove the samples that have less than 0.5 points on English language score (see~\cite{zuluaga2022atco2}).\footnote{This score ranges from 0 to 1. The higher the score, the more probability that the ATC communication is in English. The authors of ATCO2-PL database supply this information as metadata.} Finally, we select $\sim$100\,h subset for training our system.

\subsection{Automatic Speech Recognition}
\label{subsec:asr}

\begin{table}[t]
    \centering
    \caption{\textbf{Comparison of word error rates (WER) in percentages for AM and LM trained with different ATC subsets described in Section~\ref{sec:datasets}}. (i) Model 1- all data except ATCO2-PL (190\,h), (ii) Model 2- trained with public datasets (LDC-ATCC + UWB-ATCC = 33\,h), and (iii) Model 3- only ATCO2-PL (100\,h). Each model is evaluated separately on HAAWAII, LDC-ATCC and \textit{ATCO2 test set corpus}. WERs in bold denotes the top performance per test set.}
    \label{tab:asr_results}
    \scalebox{1.2}{    
    \begin{tabular}{l ccc}
        \toprule
        \rowcolor{Gray} \textbf{Model} & \multicolumn{3}{c}{\textbf{Test sets WER (\%)}} \\
        \cline{2-4}
         \rule{0pt}{2ex} & \textbf{HAAWAII} & \textbf{LDC} & \textbf{ATCO2} \\
         \midrule
        \texttt{Model 1} & \textbf{10.3} & \textbf{13.5} & 36.6 \\
        \texttt{Model 2} & 39.8 & 14.3 & 44.8 \\
        \texttt{Model 3} & 38.1 & 58.3 & \textbf{29.6} \\
        \bottomrule
    \end{tabular}
    }
\end{table}

In this work, we present results for hybrid-based ASR systems trained with the corpora listed in Table~\ref{tab:databases}.

\textbf{Experimental details:} in our experiments, conventional biphone Convolutional Neural Network (CNN)~\cite{lecun1995convolutional} + TDNN-F~\cite{povey2018semi} based acoustic models trained with Kaldi~\cite{povey2011kaldi} toolkit (i.e., nnet3 model architecture) are used. AMs are trained with the LF-MMI training framework, considered to produce state-of-the-art performance for hybrid ASR. In all the experiments, 3-fold speed perturbation with MFCCs and i-vectors features are used. Language model (LM) is trained as a statistical 3-gram model using the manual transcripts. The results are presented for the AM and LM trained with the following 3 scenarios: (i) all data except the ATCO2 dataset, (ii) public dataset (ATCOSIM + LDC-ATCC + UWB-ATCC), and (iii) 100\,h of ATCO2-PL data. We tag these models in Table~\ref{tab:asr_results} as \texttt{Model-1}, \texttt{Model-2} and \texttt{Model-3}, respectively.

\textbf{Results:} to show the performance of the ASR on various subsets of data described in Section~\ref{sec:datasets}, we report results of the ASR on 3 test sets, a private data set and 2 public datasets: (i) HAAWAII, (ii) LDC, and (iii) ATCO2 test set corpus. Each test set is a combination of both ATCo and pilot speech.
As shown in Table~\ref{tab:asr_results}, the best performance for each test set is reached when the ASR is trained with the respective data. For reproducible experiments, ASR trained with open-source data can be used.
An error-resilient ASR is critical in the ATM domain. In ~\cite{helmke2021measuring}, the authors show that achieving a lower WER increases the command recognition performance. The results in Table~\ref{tab:asr_results} also show that ATC communications in different environments result in various performances. Thus, depending on the condition of the environment, a suitable model can be used. E.g., evaluating the ASR on the ATCo subset of the HAAWAII test set results in a WER of $\sim$4\% which implies that 4 out of 100 words of an ATCo are wrong, which is acceptable in practical applications.


\subsection{High-level ATC-Entity Parser}
\label{subsec:bert_parser}

\begin{figure}[t]
  \centering
  \includegraphics[width=0.99\linewidth]{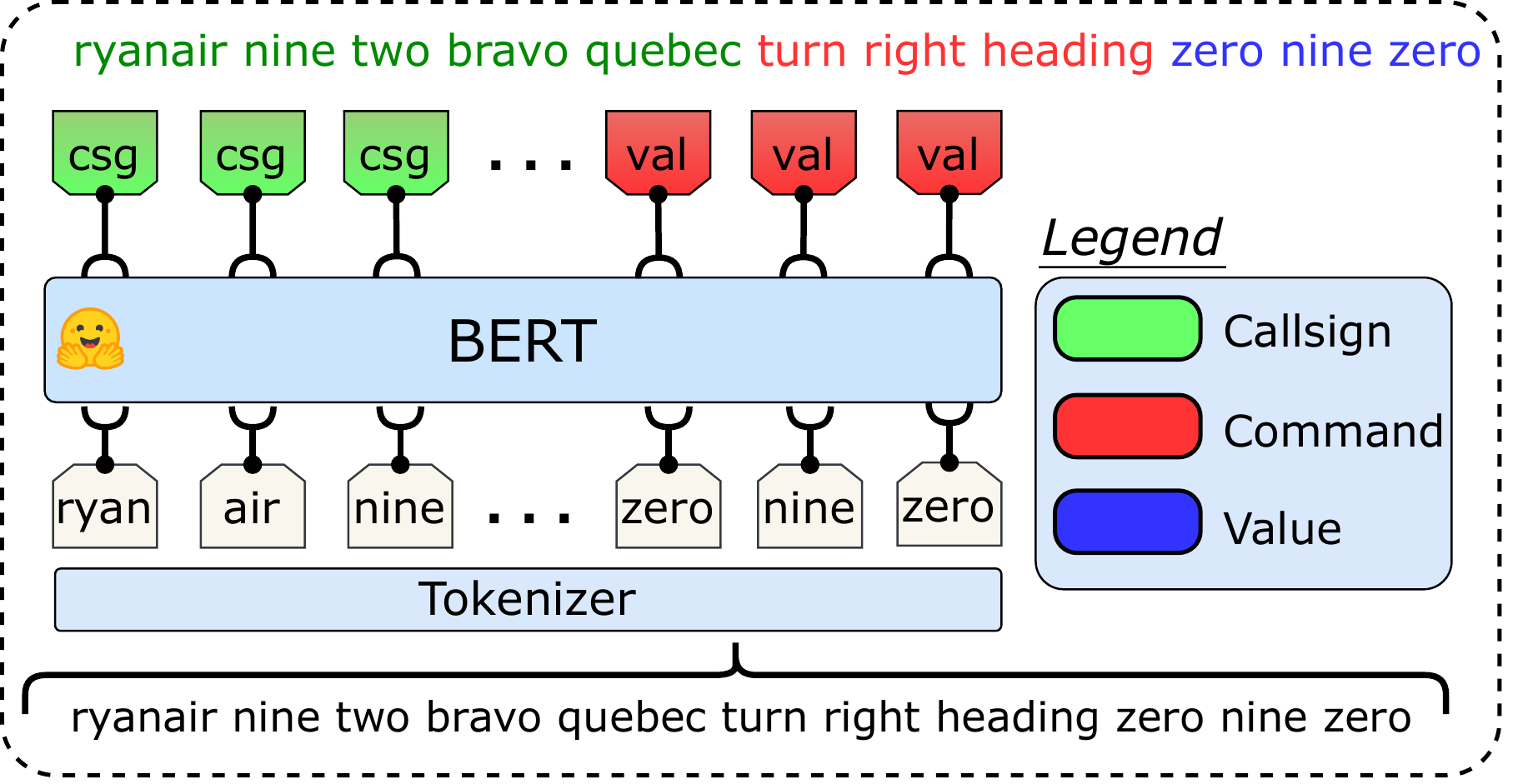}
  \caption{\textbf{High-level ATC-entity parser system}. This submodule is based on a pretrained BERT language model fetched from HuggingFace platform~\cite{wolf2020transformers}. We fine-tune BERT in the named entity recognition task for identification of key entities in transcribed ATC communications.}
  \label{fig:ner_pipeline}
\end{figure}

Air traffic control communications follow a structured grammar and set of rules. Thus, it is logical to expect that some words and phases carry special meaning in the communication. Normally, these words uttered by ATCos are translated to some actions by pilots.
We categorize three classes of entities following the \textit{ATCO2 test set corpus} ~\cite{kocour2021automatic,zuluaga2020automatic,rigault2022legal}, i.e., callsigns, commands and values. 
We developed a baseline system for recognition of these entities based on NER, as depicted in Figure~\ref{fig:ner_pipeline}. We name this model: \textit{High-level ATC-Entity Parser}. An early implementation of this system was covered in~\cite{nigmatulina2022two}. However, the authors only focus on private databases.

\textbf{Datasets}: we use the ATCO2 corpus for experiments. We split the initial subset into an 80/20 ratio to train and test our system. We did not use the other datasets from Table~\ref{tab:databases} because none of them contain gold annotations (i.e., annotations at word level) of high-level ATC-related entities. 

\textbf{Experimental details:} a NER system is trained that parses text\footnote{In our case, it parses transcripts generated by our ASR system or ground truth annotations as a proof-of-concept.} into high-level entities relevant to ATC communications.
The NER module is depicted in Figure~\ref{fig:ner_pipeline}. First, a BERT\footnote{The pre-trained version of \texttt{BERT-base-uncased} with 110 million parameters is used. URL: \texttt{\url{https://huggingface.co/bert-base-uncased}}.}~\cite{devlin-etal-2019-bert} model is downloaded from HuggingFace~\cite{wolf2020transformers,lhoest2021datasets} which is then fine-tuned on the NER task with 3k sentences ($\sim$3 hours of speech) using the \textit{ATCO2 test set corpus}, where each word has a tag. The final layer of the BERT model is replaced by a linear layer with a dimension of 8 (following the classes structure from Section 3.3 of~\cite{zuluaga2021bertraffic}, two outputs for each class). As only 3k sentences are used, a 5-fold cross-validation is conducted.
Further details about experimentation are covered in~\cite{nigmatulina2022two}. We redirected the reader to the public and open-source GitHub repository of the ATCO2 corpus (\texttt{\url{https://github.com/idiap/atco2-corpus}}). Here, the authors released Python scripts to replicate part of the results presented in this work. 


\textbf{Results:} Table~\ref{tab:ner_results} shows the results on ATCO2 test set. For each class, the results are presented using the precision, recall and F1-score metrics. The BERT-based model yielded an average of 0.97/0.82/0.87 F1-score for callsign, commands and values, respectively. We observe that the command class is the most challenging among all the classes, as they carry extra complexity compared to values and callsigns. For instance, a value is mainly composed of cardinal numbers (e.g., one, one hundred, one thousand) and some additional words, e.g., flight level. While a callsign is composed of an airline designator along with numbers and radiotelephony alphabet~\cite{allclear}. Overall, the recognition rates are above 80 points in all metrics (see Table~\ref{tab:ner_results}). However, in practical terms, further experimentation and validation needs to be undertaken with in real-life scenarios, in order to determine the minimum required performance to make the system viable. Similarly, there is still space for improvement, for instance, by adding real-time surveillance data into the system. An example is covered in~\cite{nigmatulina2022two}. Also, we should test the performance of the ATC-entity parser on top of ASR output rather than ground truths.

\begin{table}[t]
    \centering
    \caption{\textbf{Mean and Standard deviation (Std) of different performance metrics for each entity from our NER system}. Precision (indicates how accurate the model is at what it predicted), recall (how accurate the model is at identifying the correct class), and F1-score metrics are used to evaluate our high-level ATC-entity parser system trained and evaluated on ATCO2 test set.
    A 5-fold cross-validation scheme is carried out, as the train and test split are small. Results are presented for each class, i.e., callsign, command, and value. Note that the results are obtained on top of gold annotations.}
    \label{tab:ner_results}
    \scalebox{0.99}{
    \begin{tabular}{lc ccc}
        \toprule
        \rowcolor{Gray} \textbf{Class} & \textbf{Nb. Tokens} &  \multicolumn{3}{c}{\textbf{Metrics: Mean (Std)}} \\
        \cline{3-5}
         \rule{0pt}{3ex} & \textbf{Mean (Std)} & \textbf{Precision} & \textbf{Recall} & \textbf{F1-score} \\
         \midrule
        Callsign & 2274 (45) & 0.97 (0.004) & 0.98 (0.004) & 0.97 (0.005) \\
        Command & 1030 (~7) & 0.80 (0.016) & 0.83 (0.023) & 0.82 (0.020) \\
        Values & 2175 (49) & 0.86 (0.008) & 0.88 (0.015) & 0.87 (0.006) \\
        \bottomrule
    \end{tabular}
    }
\end{table}

\subsection{Repetition Generator}

The repetition generator (RG) is the core of the pseudo-pilot agent. It receives the output of the NER system, which contains the callsign, commands, and values uttered by the ATCo, and it produces a spoken response. The response is a WAV file that is played back over the ATCos' trainee headphones. It can also be stored along with the metadata for future control and assessment. In essence, this response matches the grammar of what a typical pseudo-pilot (or pilot) would reply based on the initial commands issued by the ATCo. In addition, the RG system comprises three submodules: a grammar converter, a word fixer, and a text-to-speech module (also known as a speech synthesizer). An overview of the RG system is in the red box of Figure~\ref{fig:full_pipeline}. 

\textit{Grammar converter module}: is built based on the fact that in ATC communications, pilots typically mention the callsign at the end, while ATCos do at the beginning. This module swaps the order of the entities detected by our NER system (see Section~\ref{subsec:bert_parser}) to match the pilots' `grammar' structure. 

\textit{Word fixer module}: it modifies the commands to match what a pseudo-pilot should reply.\footnote{Based on the type of communication requirements, a similar approach can be deployed to update the response. 
e.g., `create' a desirable read-back error. This can be useful in training ATCos to spot these errors. One example: \mbox{\texttt{turn right} $\rightarrow$ \texttt{turn left}}.} We perform this by applying some mapping rules, e.g., \mbox{descend $\rightarrow$ descending} or \mbox{turn $\rightarrow$ heading}. These rules ensure that the reply generated by the RG is as close as possible to standard ICAO phraseology. 
Our current word fixer module contains a list of 15 commands.
Adding additional mapping rules to a \texttt{rules.txt} file can easily update this module, allowing the system to work in different environments, e.g., ground tower or control approach. 



\textit{Text-to-Speech module:} finally, when the final textual prompt is assembled, an out-of-the-box TTS system converts the generated prompt into spoken format.

\textbf{Experimental details:} a recently proposed
non-autoregressive speech synthesizer, FastSpeech2 model~\cite{ren2020fastspeech} is used in our experiments. A brief description of the details and implementation is in Section~\ref{subsec:tts}. We download and use out-of-the-box a well-known pre-trained TTS model. Specifically, we use FastSpeech2 model from HuggingFace hub~\cite{wolf2020transformers}, which is available at the following link: \texttt{\url{https://huggingface.co/facebook/fastspeech2-en-ljspeech}}.
We use this model out-of-the-box by simply performing inference with the sentence produced by the repetition generator, i.e., the prompt of the pseudo-pilot.
Additional models, such as Tacotron~\cite{Wang2017tacotron} or Tacotron2~\cite{shen2018tacotron2},\footnote{Tacotron2 model is public and free to access in the HuggingFace hub in the following link: \texttt{\url{https://huggingface.co/speechbrain/tts-tacotron2-ljspeech}}.} can be also fine-tuned and deployed for the ATC data.

\textbf{System analysis:} during the experiments, we found out that the model can handle complicated word sequences\footnote{For example, pilot's read backs that include more than one command and values.}, which are common in ATC. However, we did not perform any qualitative analysis of the produced voice/speech by the TTS. We leave this field as future line of work. Also, one can anticipate that akin to the NER system, the TTS module can be fine-tuned on the specific field, i.e., ATC. We did not explore this area, as the main idea of the paper was to implement a simple, yet efficient pseudo-pilot system with already available open-source models. Adapting the TTS system with ATC audio is also a future line of work.

\section{Conclusions and Future Work}
\label{sec:conclusion}

In this work, we investigated an approach to generate a pseudo-pilot agent using speech and natural language processing techniques that are developed and implemented efficiently. The main modules comprise: transcribing ATCo communication with an ASR, parsing the ATC-related entities from the transcript with a NER module, and rendering a pilot-alike response using a simple repetition generator module. The ASR experimental results are presented for three different training scenarios: (i) using all available ATC data, (ii) using only open-source data, and (iii) using open-source but noisy data.
These ASR models are then evaluated on three different test sets. The results show that each test set performs better when the data is seen during training, and it indicates that ASR is essential as the correct transcript is required to generate a pilot response. The high-level entity parser tags the sequence of words to callsigns, commands, and values are trained and evaluated on the ATCO2 test set and has F1-scores above 0.80 for all the classes. And, as the primary goal in this work is to generate a simple and efficient pseudo-pilot, we deploy a rule-based dialogue system along with a generic state-of-the-art TTS system to generate a pseudo-pilot agent. 

We investigated hybrid-based systems for ASR, which can further be improved by incorporating (i) surveillance data as an additional modality and (ii) end-to-end training techniques. As mentioned earlier, the repetition generator uses a simple grammar converter and a pre-trained TTS system. As part of our future work, we consider investigating the grammar converter to (i) include greetings and complex responses and (ii) based on the surveillance data, to initiate a request from a pilot and TTS system to fine-tune it to the ATC domain. We also consider developing techniques to have a quantitative metric for evaluating the system.

\section*{Acknowledgements}
The work was supported by SESAR EC project No. 884287\,-\,HAAWAII (Highly automated air-traffic controller workstations with artificial intelligence integration). The work was also partially supported by the European Union's Horizon 2020 project No. 864702\,-\,ATCO2 (Automatic collection and processing of voice data from air-traffic communications), which is a part of Clean Sky Joint Undertaking.

\bibliographystyle{IEEEtran}
\bibliography{references}

\end{document}